\documentclass{ecai} 

\usepackage{latexsym}
\usepackage{amssymb}
\usepackage{amsmath}
\usepackage{amsthm}
\usepackage{booktabs}
\usepackage{enumitem}
\usepackage{graphicx}
\usepackage{color}
\usepackage{multirow}
\usepackage[dvipsnames]{xcolor}
\usepackage{hyperref}
\hypersetup{
    colorlinks=true,
    linkcolor=blue,
    filecolor=magenta,      
    urlcolor=blue,
}

\newcommand{\BibTeX}{B\kern-.05em{\sc i\kern-.025em b}\kern-.08em\TeX}

\begin{document}


\begin{frontmatter}


\paperid{7133} 


\title{FTIO: Frequent Temporally Integrated Objects}




\author[A]{\fnms{Mohammad}~\snm{Mohammadzadeh Kalati}\thanks{Corresponding Author. Email: mohammad.kalati@usask.ca.}}
\author[B]{\fnms{Farhad}~\snm{Maleki}}
\author[A]{\fnms{Ian}~\snm{McQuillan}} 

\address[A]{Department of Computer Science, University of Saskatchewan}
\address[B]{Department of Computer Science, University of Calgary}


\begin{abstract}
Predicting and tracking objects in real-world scenarios is a critical challenge in Video Object Segmentation (VOS) tasks. Unsupervised VOS (UVOS) has the additional challenge of finding an initial segmentation of salient objects, which affects the entire process and keeps a permanent uncertainty about the object proposals. Moreover, deformation and fast motion can lead to temporal inconsistencies. To address these problems, we propose Frequent Temporally Integrated Objects (FTIO), a post-processing framework with two key components. First, we introduce a combined criterion to improve object selection, mitigating failures common in UVOS—particularly when objects are small or structurally complex—by extracting frequently appearing salient objects. Second, we present a three-stage method to correct temporal inconsistencies by integrating missing object mask regions. Experimental results demonstrate that FTIO achieves state-of-the-art performance in multi-object UVOS. Code is available at: \textit{\url{https://github.com/MohammadMohammadzadehKalati/FTIO}}.
\end{abstract}

\end{frontmatter}

\section{Introduction and Related Works}
\label{sec:intro}
The purpose of Video Object Segmentation (VOS) is to identify the pixels belonging to each object of interest across all frames in a video sequence. This serves as a valuable pre-processing step for many real-world applications such as visual surveillance, action recognition, video summarization, and video editing~\cite{Gao01a,Perazzi_2016_CVPR}. Early VOS methods~\cite{Wang_2015_CVPR,Zhang_2013_CVPR,Papazoglou_2013_ICCV} used hand-crafted low-level features and heuristics related to foreground detection~\cite{Pei01,Gao01a}. These approaches were commonly developed and evaluated on datasets such as BMS series~\cite{Brox01,Ochs01a} and SegTack series~\cite{Tsai01a,Li_2013_ICCV}, which are limited in diversity, challenge coverage, and video length~\cite{Gao01a}. Consequently, they fall short when applied to complex datasets such as DAVIS~\cite{Perazzi_2016_CVPR} and Youtube-VOS ~\cite{Xu01,Xu02}. As a result, these early approaches are increasingly being replaced by recent deep learning and attention-based approaches, which have demonstrated significant improvements~\cite{Cheng01,Cheng_2023_ICCV,Yan_2023_CVPR,Cheng_2024_CVPR}.

In a VOS task, if annotations are provided in at least one frame during evaluation, the task is referred to as Semi-supervised VOS (SVOS). If no annotations are provided, it is called Unsupervised VOS (UVOS). Although SVOS methods generally yield more accurate segmentations, providing such annotations is time-intensive~\cite{Delatolas_2024_WACV} and prevents fully automatic segmentation. Therefore, improving UVOS approaches is critical for many real-world applications.

Several key works have recently advanced UVOS. Dual Prototype Attention (DPA)~\cite{Cho_2024_CVPR} introduces two novel prototype-based attention mechanisms, Inter-Modality Attention (IMA) and Inter-Frame Attention (IFA), to enhance UVOS. Leveraging transformers and attention mechanism, Isomer~\cite{Yuan_2023_ICCV} applies two transformer variants---Context-Sharing Transformer (CST) and Semantic Gathering-Scattering Transformer (SGST)---for low-level and high-level feature fusions, respectively, formulating a level-isomerous transformer framework for Zero-shot (Unsupervised) VOS tasks. The Hierarchical Feature Alignment Network (HFAN)~\cite{Pei01} introduces a feature alignment (FAM) module and a feature adaptation (FAT) module to process the appearance and motion features hierarchically. In another recent work~\cite{Song01a}, Generalizable Fourier Augmentation (GFA) was proposed to improve the generalizability of the developed models. To introduce a high-performing online method for UVOS, Online Adversarial Self-Tuning (OAST)~\cite{Su_2023_ICCV} integrates offline training with online fine-tuning in a unified framework.

These key works have shown promising results in UVOS. However, similar to most earlier UVOS methods~\cite{Luiten_2020_WACV, Zhuo01a}, they focus solely on producing binary segmentation masks to distinguish the foreground objects from the background. However, in many real-world applications, it is crucial to be able to differentiate between multiple object instances~\cite{Gao01a}. The Unsupervised Multi-Object Segmentation Challenge~\cite{Caelles01a,Pont-Tuset01a} was introduced to accelerate progress toward practical applications, leading to multi-object UVOS approaches~\cite{Luiten_2020_WACV, Zhou_2021_CVPR}. These methods improved both the accuracy and practical utility of UVOS by integrating instance-level segmentation modules~\cite{Gao01a}.

Analyzing high-performing multi-object UVOS approaches~\cite{Luiten_2020_WACV,Lin_2021_ICCV,Cheng_2023_ICCV} is essential for improving UVOS  methods. Propose-Reduce~\cite{Lin_2021_ICCV} was mainly designed for Video Instance Segmentation~\cite{Yang_2019_ICCV}, but it can be applied for UVOS by modifying object categories. UnOVOST~\cite{Luiten_2020_WACV} and Propose-Reduce both use Mask R-CNN~\cite{He01} to produce image-level segmentations for video frames. Mask R-CNN provides segmentations for specific categories. This limits the UVOS methods to propose only the objects within those categories. To surpass this limitation, decoupled video segmentation (DEVA)~\cite{Cheng_2023_ICCV,Cheng01a} uses an open-world image-level segmentation approach called EntitySeg~\cite{Qi01a}. This is one of the key factors that position this UVOS method as the current state-of-the-art approach applied to the Unsupervised DAVIS 2017 dataset~\cite{DAVIS_Challenge,DAVIS2017val_Benchmark,DAVIS2017testdev_Benchmark}. Although having category-agnostic segmentations is beneficial, it makes the selection of salient objects more difficult because of two reasons: (1) Category-agnostic segmentation methods include many object proposals to cover all existing objects. However, multi-object UVOS datasets set a specific limit on the number of object proposals. For example, the DAVIS Unsupervised Multi-Object Segmentation Challenge~\cite{Caelles01a} sets a limit of 20 object proposals. This limitation ensures that the applied methods only propose objects that they find the most salient. (2) The confidence score of each object proposal is less reliable in comparison to category-specific approaches, making the saliency scores based on confidence (such as the one used in UnOVOST) inefficient.

Aside from the object selection issues, challenges such as motion ambiguity, occlusions and reappearing objects, cluttered or dense environments, and appearance changes over time lead to temporal consistency issues in the predictions of UVOS approaches. As a common issue in temporal consistency, even when using state-of-the-art approaches, parts of the predicted mask of some objects get removed in some frames due to either being similar to parts of another object, over-segmentation, or severe deformations.

In this research, we propose two post-processing techniques. First, we enhance the object selection scheme of DEVA by prioritizing objects that consistently appear throughout the video sequence to improve the selection of more salient objects in video sequences with multiple objects. Next, we propose a 3-stage post-processing strategy to further enhance temporal consistency by (1) detecting potentially inconsistent frames. (2) refining the detection phase by removing frames that are mistakenly predicted to be inconsistent, and (3) utilizing image registration to restore missing object mask regions in inconsistent frames and their adjacent frames.

By addressing both object selection and temporal stability, our method substantially enhances the robustness of UVOS predictions. The primary contributions of this research are threefold: (1) We refine the selection mechanism of DEVA by prioritizing persistently visible objects, thereby improving the quality of multi-object segmentation. (2) We introduce a novel three-stage refinement strategy that effectively mitigates temporal inconsistencies in object masks across frames. (3) Our approach outperforms existing UVOS methods, achieving a $\mathcal{J}\&\mathcal{F}$ score of 75.9\% on the validation set and 67.7\% on the test set of the DAVIS dataset. It is worth mentioning that among the currently popular VOS datasets, including the DAVIS series and YouTube-VOS series that are used the most frequently~\cite{Gao01a} and the new challenging MOSE dataset~\cite{Ding_2023_ICCV}, only DAVIS 2017 has a specific dataset for multi-object UVOS.
\section{Data and Methodology}
\label{sec:method}
\noindent\textbf{Dataset:} To evaluate our method, we use the Unsupervised DAVIS dataset, which consists of 60 video sequences in the training set, 30 sequences in validation set, 30 sequences in the test-dev set, and 30 sequences in the test-challenge set~\cite{Caelles01a}. The test-challenge video sequences were designed for the DAVIS Challenge~\cite{Caelles01a} and are no longer available. The challenge is also closed. Therefore, we report our results on the validation and test-dev sets. The DAVIS dataset is specifically designed for multi-object VOS, and its ground-truth annotations primarily include objects that would mostly capture human attention when watching the whole video sequence~\cite{Caelles01a}.

\noindent\textbf{Evaluation Metrics:} We assess the proposed approach using three evaluation measures most commonly used for VOS tasks. Jaccard Index $\mathcal{J}$ for region similarity, F score $\mathcal{F}$ for boundary accuracy, and their average $\mathcal{J\&F}$. These evaluation measures are defined as follows:
\begin{equation}
\mathcal{J} = \frac{| M \cap G |}{| M \cup G |},\
\mathcal{F} = \frac{2P_cR_c}{P_c + R_c},\
\mathcal{J}\&\mathcal{F} = \frac{(\mathcal{J} + \mathcal{F})}{2}
\end{equation}
where $G$ and $M$ refer to the ground-truth and predicted segmentation mask, respectively. Also, $P_c$ and $R_c$ are precision and recall computed from the points in contours $c(M)$ and $c(G)$~\cite{Gao01a}. We report all results using evaluation measures expressed as percentages.

\noindent\textbf{Baseline \& Implementation Details:}
Our post-processing approach can be applied to UVOS methods and it is especially effective on methods designed for multi-object UVOS. To demonstrate the potential of our approach and provide meaningful comparisons with other UVOS methods, we use the decoupled video segmentation (DEVA) method introduced in~\cite{Cheng_2023_ICCV,Cheng01a} for UVOS as the baseline of our method. DEVA was introduced in 2023 and is currently the state-of-the-art for multi-object UVOS. It uses the results of EntitySeg~\cite{Qi01a} to provide the image-level segmentations. Then, it feeds these segmentations with the original frames to a temporal propagation model, which is a modified version of XMem~\cite{Cheng01} designed for SVOS. We use the pre-trained model provided by DEVA for UVOS and assess our proposed post-processing approach. All experiments were conducted using a single NVIDIA L40 GPU.

\subsection{Frequent Temporally Integrated Objects (FTIO)}
\label{sec:ftio}

\begin{figure*}
\centering
  \includegraphics[width=0.9\textwidth]{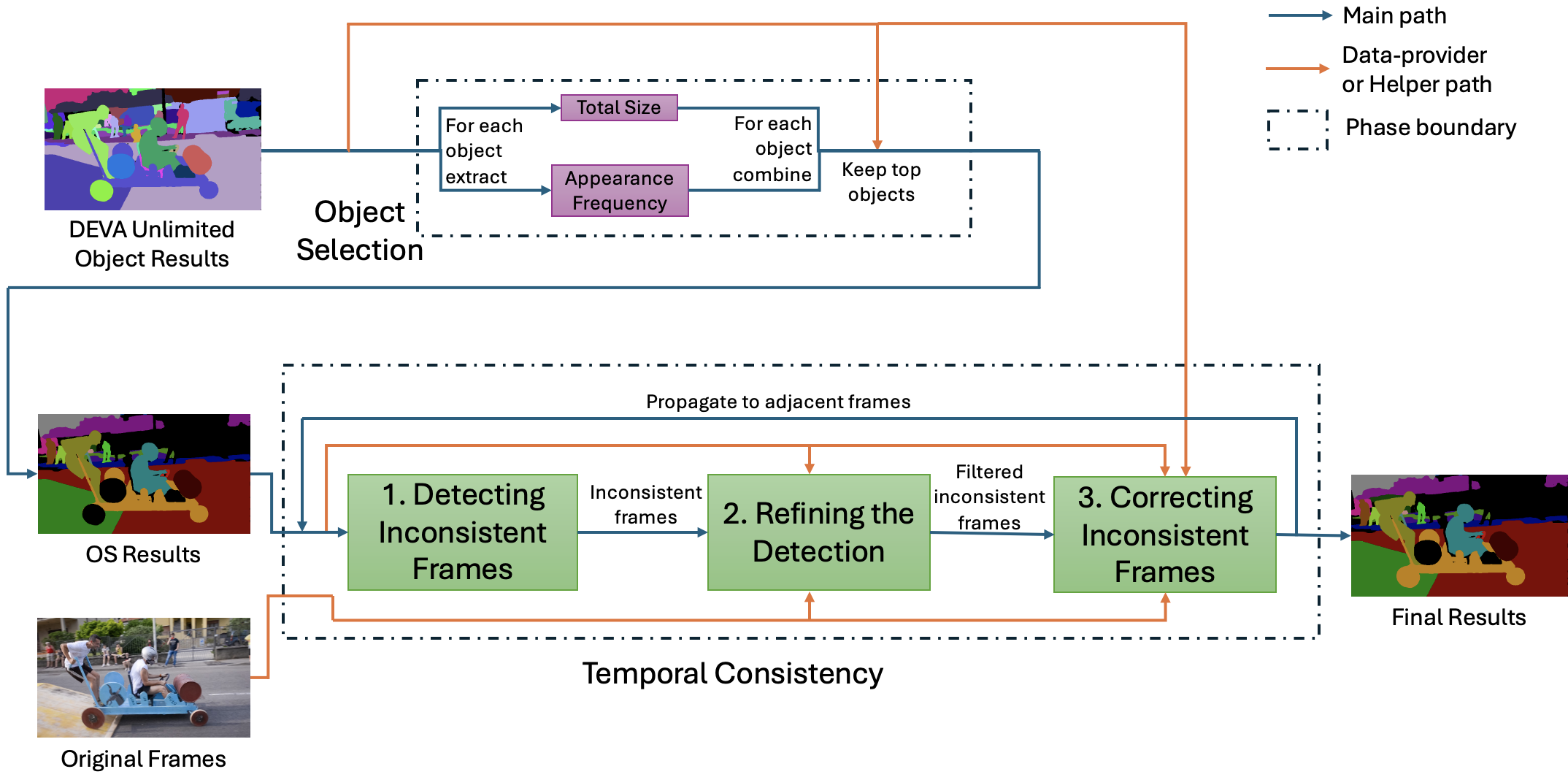}
  \caption{The overall pipeline of our FTIO approach. First, in the Object Selection (OS) phase, the total size and appearance count of the masks of each object are extracted and combined to rank the objects and keep masks of the top ones. Then, in the Temporal Consistency (TC) phase, the inconsistent frames (frames with inconsistent object masks) in OS results are detected, the detection is refined, and the inconsistent object masks are corrected. This phase is propagated to adjacent frames and repeated until all detected inconsistencies are resolved or until the number of repetitions equals the number of frames in the video sequence. The ``DEVA Unlimited Object Results'', ``OS Results'', and ``Final Results'' images are obtained by reproducing the results of DEVA~\cite{Cheng01a} before limiting the number of objects, applying OS phase of the FTIO approach to DEVA~\cite{Cheng01a}, and applying the FTIO approach entirely to DEVA~\cite{Cheng01a}, respectively, using the code provided in the Github repository of DEVA~\cite{Cheng01a}~\cite{DEVA_Github} (licensed under CC BY-NC-SA 4.0). The source of the “Original Frames” image is from the DAVIS dataset~\cite{Perazzi_2016_CVPR} (licensed under CC BY-NC 4.0).}
  \vspace{+1em}
  \label{fig:pipeline}
\end{figure*}
Figure~\ref{fig:pipeline} illustrates the overall pipeline for the proposed approach. The inputs consist of original frames and all object masks predicted by a UVOS method, while the outputs are frames with refined object masks.  The proposed pipeline has two phases: Object Selection (OS) and Temporal Consistency (TC). 

At the OS phase, we compute the appearance count of each object in the UVOS mask frames of each video sequence. We combine this with the size of the masks of the corresponding objects in the UVOS mask frames to rank the object proposals. The size of the mask of each object is determined by adding the area of all masks belonging to that object over all frames. The top ranked object mask proposals are the output of this step. Then, TC phase is performed for multiple iterations. TC phase has three stages. First, a sliding window is iterated over the results of OS. Investigating the changes in the area of masks of each object within a window of frames to detect inconsistent frames for each object. The detected frames are then filtered using spatial information in the second stage to correctly ignore the occluded but not inconsistent frames. After that, the last stage uses image registration to map the frames with temporally consistent object masks to the frames in which parts of some object masks are missing. Finally, we use this mapping to correct these missing parts and complement our mapping using the UVOS mask frames. We provide a detailed discussion of each post-processing phase in Sections~\ref{sec:os}~and~\ref{sec:tc}.

\noindent\textbf{Notation:}
At time $t$, we denote the corresponding elements as follows: original frame as $F_t$, unlimited UVOS method predictions as $D_t$, the OS result as ${OS}_t$, and the result of TC as an entire module as ${TC}_t$. If a pair of time indices, such as $t$ and $t'$, is used, we replace the single time index with $[t,t']$. Applying each phase or function to an input is represented with parentheses. For example, ${OS}(F_t)$ denotes applying the OS phase to the original frame at time $t$. To refer to the mask of an object in a specific frame (time), we add a comma followed by the index of the object to the subscript. For instance, ${TC}_{t,m}$ represents the mask of an object with index $m$ at the TC result at time $t$.
\subsubsection{Object Selection}
\label{sec:os}
\begin{figure}
\centering
\includegraphics[width=\linewidth]{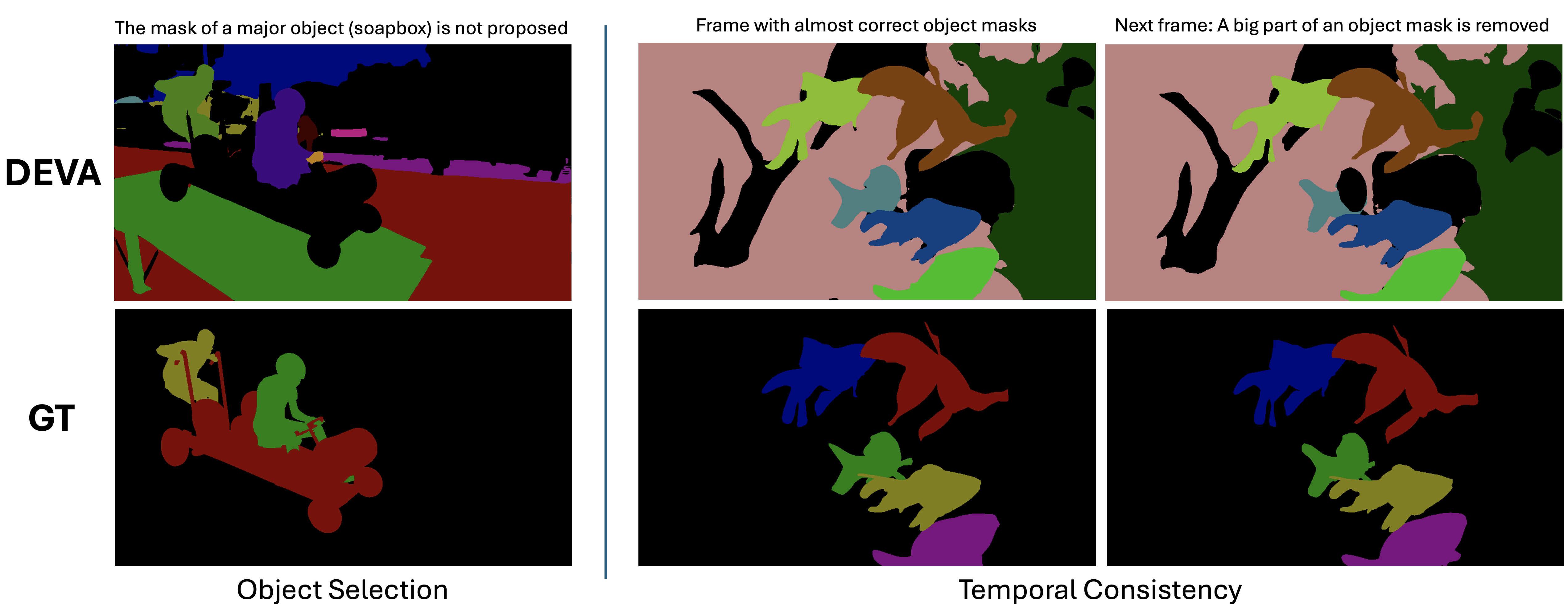}
\caption{Example of defects in object selection and temporal consistency in the predicted mask frames of a UVOS method. The images in the ``DEVA'' row are obtained by reproducing the results of DEVA~\cite{Cheng01a} using the code provided in the Github repository of DEVA~\cite{Cheng01a}~\cite{DEVA_Github} (licensed under CC BY-NC-SA 4.0). The source of ground-truth images (in the ``GT'' row) in this figure is from the multi-object unsupervised DAVIS dataset~\cite{Perazzi_2016_CVPR,Pont-Tuset01a,Caelles01a} (licensed under CC BY-NC 4.0).}
\vspace{+1em}
\label{fig:deva_issues}
\end{figure}

A common assumption in many object tracking applications is that the target objects are present in the first frame, providing a reliable reference for subsequent tracking. This assumption also holds for the DAVIS dataset~\cite{Caelles01a}, where all intended objects appear in the first frame of each video sequence. Consequently, one might consider selecting the most salient objects by focusing on the first frame.
However, this approach is not always effective, as the image-level segmentation process may fail to propose certain objects in the first frame. For instance, Figure~\ref{fig:deva_issues} illustrates a case in which DEVA fails to predict the mask of a major object (soapbox) because its image-level mask was not detected in the first frame.

To address this limitation, rather than selecting only the largest proposed objects, we incorporate an additional criterion that accounts for objects that consistently appear throughout a video~\cite{Caelles01a}. Specifically, we adopt a combination of two criteria to enhance the robustness and reliability of the object selection process. We provide a detailed explanation of this approach below.

\noindent\textbf{Object Size:}
To eliminate small background segments, which are often the result of noise in the initial predictions, we prioritize objects with larger mask areas, as they are less likely to be affected by noise~\cite{Cheng_2023_ICCV,Cheng01a}. To achieve this, we compute the size of an object $A$ in a video sequence with $n$ frames as follows:
\begin{equation}
\label{eq:object_size}
S_A = \sum_{i=1}^{n}\frac{area(D_{i,A})}{area(D_{i})}
\end{equation}
The $area$ function calculates the number of pixels in either a frame, mask, or region (specified by its operand). So, in equation \ref{eq:object_size}, $area(D_{i})$ and $area(D_{i,A})$ represent the number of pixels in frame $i$ and in the mask of object $A$ in frame $i$, respectively.

Our approach does not rely solely on size, as some salient objects might be relatively small. For instance, in comparison to a football pitch, players are small but they are typically considered salient.

\noindent\textbf{Object Mask Appearance Count:}
Salient objects are expected to consistently appear across multiple consecutive frames in a video sequence~\cite{Caelles01a}. To identify the most frequently occurring object masks, we rank all proposed objects in the sequence based on the number of UVOS mask frames that include the mask of each object. The object mask appearance count for an object $A$ is defined as follows:
\begin{equation}
\label{eq:object_mask_count}
N_A = \sum_{i=1}^{n}appear(D_{i,A})
\end{equation}
The $appear$ function equals $1$ if at least a pixel of the mask of object $A$ is present in frame $i$, otherwise it equals $0$.

\noindent\textbf{Combined criterion:}
Finally, we combine these measures using their linear combination. We fix the coefficient of $N_A$ as $1$ and use the coefficient of $S_A$, denoted by $\alpha$ to tune the trade-off between these measures. Hence:
\begin{equation}
\label{eq:comb_object_selection}
C_A = N_A + \alpha S_A
\end{equation}
Performing ablation experiments, we find $\alpha = 5$ appropriate for this hyper-parameter. Then, we extract the objects with top $C$ values (top 20 for DAVIS dataset~\cite{Caelles01a}) and keep only the masks of these objects as the result of the OS phase.
\subsubsection{Temporal Consistency}
\label{sec:tc}
As discussed in Section \ref{sec:intro}, UVOS methods---particularly those relying on image-level segmentations---may remove parts of the predicted masks of an object in some video frames due to  similarity with parts of other objects, over-segmentation, or severe deformations, as illustrated in Figure \ref{fig:deva_issues}. 
To alleviate this, we propose a 3-stage method to make the object masks in frames more temporally consistent.

\noindent\textbf{Detecting Inconsistent Frames:} 
At this stage, we focus only on the object mask frames produced by the OS phase, which are referred to as OS frames for conciseness. In a window of OS frames, if an object mask differs significantly from the corresponding mask in the consecutive neighboring frames, we identify it as inconsistent. To quantify this difference, we define Mask over Union ($MoU$) as a measure of inconsistency. $MoU$ of object $A$ at time $t$ in a given window is defined as the $area$ of the mask of object $A$ at $t^{\mbox{th}}$ frame, divided by the $area$ of the union of all corresponding object masks across the window. This can be formulated as:
\begin{equation}
\label{eq:mou}
\begin{split}
MoU_{t,A,t_s,|w|} & = \frac{area(OS_{t,A})}{area(\bigcup_{i=t_s}^{t_s + |w|}OS_{i,A})}
\end{split}
\end{equation}
where $|w|$ is the length of the window which starts at time $t_s$ and ends at time $t_s+|w|$. Then, we compare this $MoU$ with the standard deviation of the $area$ of the masks in the same window. Considering all windows of length $|w|$ where  $t_s \le t < t_s + |w|$, if more than $|w|/2$ times the $MoU$ of object $A$ at time $t$ is higher than the standard deviation, we identify the $t^{\mbox{th}}$ frame as an inconsistent frame for object $A$. We use $|w| = 5$ since we tend to have at least 3 signals indicating inconsistency of some OS frames to be confident about the detection. In addition, the object can change substantially in larger windows, making the $MoU$ less effective in detecting the exact inconsistent frame indices.

$MoU$ is informative in detecting inconsistency, especially when parts of an object mask are removed or labelled as another object mask. Changes in $MoU$ are also noticeable in zooming scenarios, but we distinguish them from the inconsistent frames by knowing that, although there is a continuous change in the size of the masks during zooming, the center of mass of objects remains almost fixed.

\noindent\textbf{Refining the Detection of Inconsistent Frames:}
Occlusion is a major challenge in VOS that causes noticeable changes in the object masks. To refine the detection in the previous stage and separate occluded frames from the inconsistent ones, we use the information from the original images. We can detect a significant change in the intensity of pixels among corresponding object masks in a pair of frames as (a probable) occlusion. To explain how this is performed in detail, consider $A$ as the object of interest, $t$ as the inconsistent frame index, and $t'$ as the index of the frame adjacent to the inconsistent frame (performed once for $t'=t-1$ and once for $t'=t+1$). We are interested in the regions of larger mask $L(OS_{[t,t'],A})$ which are not present in the smaller mask $S(OS_{[t,t'],A})$. To extract this region of difference $DR$ between corresponding object masks in two consecutive frames, we subtract the intersection of both object masks in OS frames from the larger mask:
\begin{equation}
\label{eq:large-small-mask-difference}
\begin{split}
DR_{[t,t'],A} = L(OS_{[t,t'],A}) - (L(OS_{[t,t'],A}) \cap S(OS_{[t,t'],A}))
\end{split}
\end{equation} 
Then, for each of these object masks, we compute a vector of pixel counts (histogram, denoted by $H$) of RGB intensities in the region of the object mask in the corresponding original frames. Next, we compute the Manhattan distance between the histograms as follows:

\begin{equation}
\label{eq:histogram-difference}
\begin{split}
MDH_{[t,t'],A} = \sum|H(DR_{[t,t'],A}(F_{t'})) - H(DR_{[t,t'],A}(F_{t}))|
\end{split}
\end{equation}
If the Manhattan distance between histograms is above a specified threshold, we identify this as occlusion and ignore the inconsistency between object masks. This threshold is determined based on the size of object masks as the smaller objects probably experience more severe deformations in consecutive frames in comparison to larger objects. Furthermore, to calculate more accurate $MDH$ for objects with fast motion, we shift one of the object masks to align the center of mass of the other object mask.

\noindent\textbf{Correcting Inconsistent Frames:} At this stage, we correct the inconsistent frames in which some parts of at least one object mask are missing. To this end, we compute optical flow $OF$ for object $A$, once between each inconsistent original frame and its previous frame, and once between each inconsistent original frame and its next frame. The time index of the inconsistent frame and its adjacent frame are denoted by $t$ and $t'$, respectively. To have a fast and robust frame registration utilizing optical flow, we use the iterative registration of local window, known as iterative Lucas-Kanade (LK)~\cite{LeBesnerais01, Lucas01}. Exploiting $OF$ for frame registration, we project the inconsistent OS frames to their adjacent OS frames and vice versa. Then, we extract the difference region $DF$ between the mask of object $A$ in the projected inconsistent OS frame and its adjacent one as follows:
\begin{equation}
\label{eq:difference-region}
\begin{split}
&UF_{[t,t'],A} = OS_{t',A} \cup OF_{[t,t'],A}(OS_{t})\\
&IF_{[t,t'],A} = OS_{t',A} \cap OF_{[t,t'],A}(OS_{t})\\
&DF_{[t,t'],A} = UF_{[t,t'],A} - IF_{[t,t'],A}
\end{split}
\end{equation}
We do the same between the inconsistent OS frame and the projected adjacent OS frame, using the negative of optical flow to perform the projection in the correct direction. We then compare the size of the masks of object $A$ in these OS frames, to determine from which of them (the inconsistent frame or its adjacent frame) parts of an object mask are missed. Next, we perform some post-processing to remove misleading connected components in the extracted missing parts. This post-processing includes removing small objects and holes from the extracted missing parts, followed by performing erosion to remove thin regions such as the boundaries of objects. After that, we add the reliable connected components that include the remaining regions to the object mask frame that is missing them. If the connected components to add are minor, we rely on the OS frames and do not add the connected components.

In addition, we can take advantage of the objects proposed by the UVOS method before limiting the number of object proposals. If the added part is a large enough fraction of another object mask in the UVOS mask frames, it is likely that an over-segmentation has occurred so we add that object mask completely to integrate the over-segmented parts of object masks. Moreover, if the inconsistent part belongs to the mask of another object in the OS frame that we aim to modify, we determine whether labeling that part as object $A$ helps histogram of intensity of pixels (used in the previous stage) become more similar between masks of object $A$ in both frames.

\noindent\textbf{Propagation to Adjacent Frames and Repairing Frames Recursively:} 
This 3-stage method integrates the missing parts of object masks in frames detected as inconsistent or in their adjacent frames. When parts of an object mask are missing in multiple consecutive frames, we apply this method recursively until no new temporally inconsistent frames remain or until TC is propagated to all frames.
\section{Experiments}
\label{sec:experiment}
\begin{figure*}
\centering
  \includegraphics[width=\textwidth]{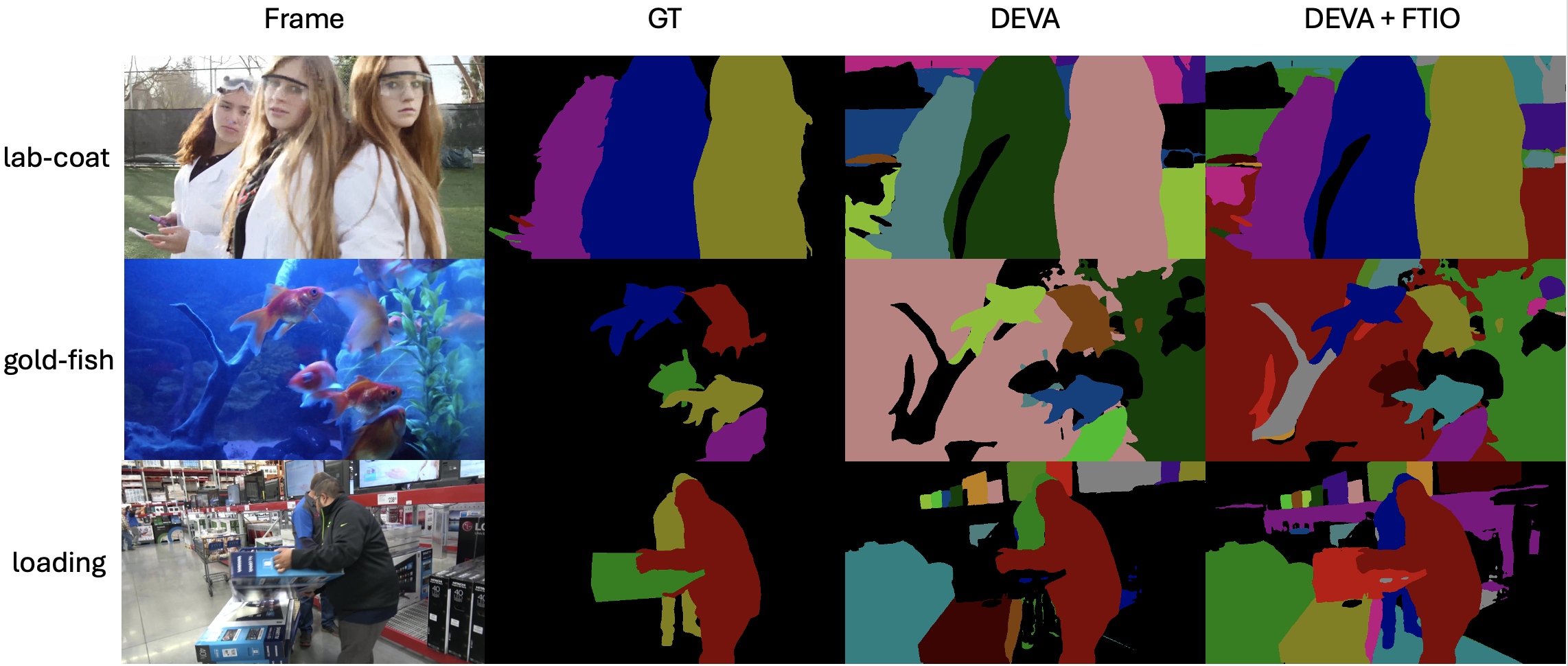}
  \caption{Qualitative Results of FTIO on samples from the DAVIS 2017 dataset. Applying FTIO leads to following improvements: Identifying segmentation masks of the phones in the hands of a person in the ``lab-coat'' video sequence. Fixing the incomplete segmentation of one of the goldfish in the ``gold-fish'' video sequence. Identifying segmentation mask of the TV box and correcting the segmentation of legs of a person in the ``loading'' video sequence. The images in the ``DEVA'' and ``DEVA + FTIO'' columns are obtained by reproducing the results of DEVA~\cite{Cheng01a} and applying the FTIO approach to DEVA~\cite{Cheng01a}, respectively, using the code provided in the Github repository of DEVA~\cite{Cheng01a}~\cite{DEVA_Github} (licensed under CC BY-NC-SA 4.0). The source of images in the ``Frame'' and ``GT'' columns in this figure is from the multi-object unsupervised DAVIS dataset~\cite{Perazzi_2016_CVPR,Pont-Tuset01a,Caelles01a} (licensed under CC BY-NC 4.0).
}
  \vspace{+1em}
  \label{fig:qualitative_results}
\end{figure*}
\noindent\textbf{Qualitative Results:} Figure \ref{fig:qualitative_results} illustrates the effect of our proposed post-processing methods for samples from the DAVIS dataset. In the ``lab-coat'' video sequence, two cell phones in the hands of a person are annotated as salient objects. DEVA proposes no masks for them since their image-level masks are not predicted in the first frame. Also, they are very small in comparison to other objects in this video sequence. Conversely, our OS method proposes masks for these objects where their masks are predicted by the image-level segmentation model. Investigating the results of the ``gold-fish'' video sequence, DEVA misses a big part of one of the goldfish after frame 60 due to the over-segmentation of the goldfish as two objects. This issue is identified and fixed by our proposed TC phase. Finally, investigating the results of the ``loading'' video sequence, DEVA does not predict the mask of the TV box, since it is not predicted in the first frame. OS predicts the TV box mask in most frames where the mask is predicted by the image-level segmentation model. Moreover, TC corrects some incomplete TV box masks and the corrections are propagated to the frames that miss the TV box masks. Hence, FTIO leads to meaningful and effective improvements in UVOS. Failure cases of OS and TC are provided in Figures \ref{fig:os_failure} and \ref{fig:tc_failure} in the supplementary materials (Section~\ref{sec:supplementary}).\par
\noindent\textbf{Quantitative Results:} Table \ref{tab:global} presents the result of the proposed approach and compares them with previously evaluated methods on the Unsupervised DAVIS 2017 dataset. As shown, our FTIO approach sets a new state-of-the-art on both (validation and test-dev) sets.
\begin{table}
  \centering
  \caption{Quantitative Results and comparisons of multi-object UVOS approaches on Unsupervised DAVIS 2017 dataset. The first, second, and third top results are indicated in \underline{\textbf{bold \& underlined}}, \textbf{only bold}, and \underline{only underlined}, respectively. For DEVA, the results in the table are reported in their paper~\cite{Cheng_2023_ICCV,Cheng01a}. Evaluating the code provided in their Github repository~\cite{DEVA_Github} results in 74.3\% $\mathcal{J}\&\mathcal{F}$ on the validation set and 62.9\% $\mathcal{J}\&\mathcal{F}$ on test-dev set. FTIO in the table indicates our FTIO post-processing approach applied to this method. As shown, our methods obtain the best results on both sets.}
  \vspace{+1.5em}
  \begin{tabular}{p{6em}|c|c|c|c|c|c}
    \hline
    \multirow{2}{*}{\textbf{Method}}& \multicolumn{3}{c|}{validation} & \multicolumn{3}{c}{test-dev}\\
    \cline{2-7}
     & $\mathcal{J}\&\mathcal{F}$ & $\mathcal{J}$ & $\mathcal{F}$ & $\mathcal{J}\&\mathcal{F}$ & $\mathcal{J}$ & $\mathcal{F}$\\
    \hline\hline
    RVOS~\cite{Ventura_2019_CVPR} & 41.2 & 36.8 & 45.7 & 22.5 & 17.7 & 27.3\\\hline
    PDB~\cite{Song_2018_ECCV} & 55.1 & 53.2 & 57.0 & 40.4 & 37.7 & 43.0\\\hline
    MuG-W~\cite{Lu_2020_CVPR} & - & - & - & 41.7 & 38.9 & 44.5\\\hline	
    AGS~\cite{Wang_2019_CVPR} & 57.5 & 55.5 & 59.5 & 45.6 & 42.1 & 49.0\\\hline
    ALBA~\cite{Gowda01} & 58.4 & 56.6 & 60.2 & - & - & -\\\hline
    MATNet~\cite{Zhou01a} & 58.6 & 56.7 & 60.4 & - & - & -\\\hline
    STEm-Seg~\cite{Athar01} & 64.7 & 61.5 & 67.8 & - & - & -\\\hline
    MAST~\cite{Lai_2020_CVPR} & 65.5 & 63.3 & 67.6 & - & - & -\\\hline
    UnOVOST~\cite{Luiten_2020_WACV} & 67.9 & 66.4 & 69.3 & 58.0 & \underline{54.0} & \underline{62.0}\\\hline
    Propose-Reduce~\cite{Lin_2021_ICCV} & 70.4 & 67.0 & 73.8 & - & - & -\\\hline
    DEVA (EntitySeg)~\cite{Cheng_2023_ICCV,Cheng01a} & \underline{73.4} & \underline{70.4} & \underline{76.4} & \underline{62.1} & - & -\\\hline
    FTIO(OS) & \textbf{75.6} & \textbf{72.8} & \textbf{78.3} & \underline{\textbf{67.7}} & \underline{\textbf{61.7}} & \underline{\textbf{73.7}}\\\hline
    FTIO & \underline{\textbf{75.9}} & \underline{\textbf{73.2}} & \underline{\textbf{78.7}} & \underline{\textbf{67.7}} & \underline{\textbf{61.7}} & \underline{\textbf{73.7}}\\
    \hline\hline
  \end{tabular}
  \label{tab:global}
\end{table}
\subsection{Ablation Studies}
\noindent\textbf{Effect of TC:} 
To investigate the effect of TC, a comparison of the results between using only OS and using both OS and TC is presented in Table \ref{tab:per-sequence}. Focusing on the ``TC Effect'' column, there is more than a 10\% increase in $\mathcal{J}$ and in $\mathcal{F}$ for the gold-fish5 and soapbox3 objects, respectively. Investigating the considerable changes, except for three objects in the ``india'' video sequence where there is a decrease in $\mathcal{J}$ or $\mathcal{F}$ due to the complex scene and misidentification of some objects, applying TC almost always leads to better results.
\begin{table}
  \centering
  \caption{Results of our FTIO approach on objects of Unsupervised DAVIS 2017 video sequences, using only OS and using both OS and TC methods. Only the objects with at least $1\%$ change by adding TC are shown. The effectiveness of TC is indicated in the TC Effect double-column. Results in \textcolor{ForestGreen}{forest green} and \textcolor{Red}{red} indicate ``increase'' and ``decrease'' in the evaluation metric when applying TC, respectively.}
  \vspace{+1.5em}
  \label{tab:per-sequence}
  \begin{tabular}{p{5em}|cc|cc|cc}
    \hline
    \multirow{2}{*}{\textbf{Object}}& \multicolumn{2}{c|}{OS + TC} & \multicolumn{2}{c|}{OS} & \multicolumn{2}{c}{TC Effect}\\
    \cline{2-7}
    &$\mathcal{J}$ & $\mathcal{F}$ & $\mathcal{J}$ & $\mathcal{F}$ & $\mathcal{J}$ & $\mathcal{F}$\\
    \hline
    gold-fish5 & 83.5 & 88.6 & 70.2 & 82.3 & \textcolor{ForestGreen}{+13.3} & \textcolor{ForestGreen}{+6.3}\\
    india4 & 17.3 & 17.3 & 19.8 & 19.9 & \textcolor{Red}{-2.5} & \textcolor{Red}{-2.6} \\
    india5 & 20.4 & 26.9 & 19.9 & 25.2 & \textcolor{ForestGreen}{+0.5} & \textcolor{ForestGreen}{+1.7} \\
    india6 & 18.7 & 24.4 & 17.4 & 28.5 & \textcolor{ForestGreen}{+1.3} & \textcolor{Red}{-4.1} \\
    india8 & 23.5 & 23.5 & 25.9 & 25.9 & \textcolor{Red}{-2.4} & \textcolor{Red}{-2.4} \\
    lab-coat5 & 24.2 & 29.2 & 23.4 & 28.1 & \textcolor{ForestGreen}{+0.8} & \textcolor{ForestGreen}{+1.1}\\
    loading2 & 91.1 & 97.8 & 86.7 & 97.1 & \textcolor{ForestGreen}{+4.4} & \textcolor{ForestGreen}{+0.7}\\
    loading3 & 69.4 & 78.3 & 63.7 & 72.8 & \textcolor{ForestGreen}{+5.7} & \textcolor{ForestGreen}{+5.5}\\
    paragliding-launch1 & 2.1 & 9.7 & 2.1 & 5.9 & 0 & \textcolor{ForestGreen}{+3.8}\\
    shooting3 & 55.2 & 53.9 & 53.4 & 52.3 & \textcolor{ForestGreen}{+1.8} & \textcolor{ForestGreen}{+1.6}\\
    soapbox3 & 46.9 & 64.8 & 39.2 & 51.6 & \textcolor{ForestGreen}{+7.7} & \textcolor{ForestGreen}{+13.2}\\
    \bottomrule
  \end{tabular}
\end{table}

\noindent\textbf{Effect of criteria used for OS:} As shown in the Single-Criterion tests in Table \ref{tab:ablation}, only using appearance count results in the worst outcomes in the validation set, while using only large objects gives slightly better results than combining appearance count and size. However, this outcome is scenario-dependent and may vary in other datasets with different properties, such as including many salient small objects. On the test-dev set, using only size results in 64.99 $\mathcal{J}\&\mathcal{F}$, while appearance count alone yields 68.14 $\mathcal{J}\&\mathcal{F}$. The results of Combined-Criterion tests with several values for $\alpha$ are compatible with the inferred pattern of Single-Criterion tests. Choosing low values for $\alpha$ (for example, $\alpha = 1$) prioritizes some noisy small regions that persistently appear in the background over the salient objects. Values higher than 5 lead to almost no considerable enhancements and they can result in less accurate results on datasets with many small salient objects, such as the players and the ball on a football pitch. $\alpha = 5$ results in an appropriate trade-off between size and appearance count. Hence, we report our final results using $\alpha = 5$.\par
\noindent\textbf{Effect of Refining step in TC:} As indicated in Table \ref{tab:ablation} for the No-Refining test, the results worsen if we ignore the Refining step as occluded objects with no inconsistency are mispredicted to be inconsistent.\par
\noindent\textbf{Effect of using object masks predicted by the baseline UVOS method:} If we do not use the object masks predicted by the baseline UVOS method in the ``Correcting Inconsistent Frames'' stage, we may fail to include the entire region missed by the object masks. This is because the frame registration performs alignment of the objects in frames approximately. The result of the Not-Use-All-Objects test in Table \ref{tab:ablation} confirms this outcome.\par
\noindent\textbf{Effect of the propagation of TC:} We performed the TC-propagation test to compare the results for different (1, 5, and 10) maximum frames, with the results provided in Table \ref{tab:ablation}. Propagating TC over more frames (up to the total number of frames in the video sequence) increases confidence in correcting all inconsistent frames.\par
\begin{table}
  \centering
  \caption{Ablation studies on our FTIO approach. We test our approach against four test categories (seven tests in total) to show the effectiveness of different stages of FTIO.}
  \vspace{+1.5em}
  \begin{tabular}{p{17em}|c|c|c}
    \hline
    Test & $\mathcal{J}\&\mathcal{F}$ & $\mathcal{J}$ & $\mathcal{F}$\\
    \hline
    Single-Criterion: appearance count (OS) & 73.87 & 71.08 & 76.66\\
    Single-Criterion: size (OS) & 76.33 & 73.63 & 79.04\\\hline
    Combined-Criterion: $\alpha$ = 1 & 74.21 & 71.40  & 77.01\\
    Combined-Criterion: $\alpha$ = 3 & 75.86 & 73.19 & 78.53\\
    Combined-Criterion: $\alpha$ = 5 & 75.94 & 73.24 & 78.65\\
    Combined-Criterion: $\alpha$ = 13 & 75.90 & 73.20 & 78.60\\
    Combined-Criterion: $\alpha$ = 50 & 75.88 & 73.16 & 78.59\\
    \hline
    No-Refining (OS + TC) & 75.85 & 73.15 & 78.55\\
    \hline
    Not-Use-All-Objects (OS + TC) & 75.76 & 72.98 & 78.55\\
    \hline
    TC-Propagation:1 & 75.58 & 72.83 & 78.32\\
    TC-Propagation:5 & 75.76 & 73.06 & 78.47\\
    TC-Propagation:10 & 75.88 & 73.17 & 78.58\\\hline
  \end{tabular}
  \label{tab:ablation}
\end{table}
\section{Limitations}
\label{sec:limitation}
As explained in Section \ref{sec:tc}, we compute optical flow and apply frame registration, to perform the ``Correcting Inconsistent Frames'' stage. These tasks are the computational bottleneck of our post-processing approach, as they are computationally intensive and we need to apply them between every inconsistent frame and its adjacent frames. In our approach, the inference time to process all frames of each video sequence is between a minute and 15 minutes, and the overall inference time on the validation set of the Unsupervised DAVIS 2017 dataset is almost 2.5 hours. The inference time of each video sequence depends on the number of frames, proposed objects, and the propagation of TC (Section \ref{sec:tc}). As a main requirement of our approach, performing optical flow and frame registration is inevitably necessary to determine the missing parts of object masks. The other alternatives, such as feature-based registration or shifting masks based on the movement of the center of mass of the object mask, are not accurate enough to satisfy this goal. Hence, this approach is proposed for tasks that require accurate and consistent segmentation in offline settings. A comparison of efficiency between our post-processing steps in FTIO (in an updated version, but not updated for efficiency) and DEVA~\cite{Cheng_2023_ICCV,Cheng01a} is provided in Table \ref{tab:inference_time_comparison} in the supplementary materials (Section~\ref{sec:supplementary}). 

To train and evaluate deep learning models, most of the recent VOS methods use DAVIS~\cite{Perazzi_2016_CVPR} and YouTube-VOS~\cite{Xu01,Xu02} datasets, because of the variety of challenges, as well as the number of videos and annotated frames in these datasets. MOSE~\cite{Ding_2023_ICCV} is also a new dataset for VOS drawing attention that contains many complex scenes. Among these popular datasets, only DAVIS 2017 has a specific dataset for multi-object UVOS. Other datasets are mainly provided for SVOS or single-object UVOS. YouTube-VIS dataset is designed for the task of Video Instance Segmentation (VIS)~\cite{Yang_2019_ICCV} which is closely related to UVOS. However, in addition to the segmentation which is almost the same in both tasks, VIS requires a classification for each object instance.
\section{Conclusion}
\label{sec:conclusion}
In this paper, we introduced Frequent Temporally Integrated Objects (FTIO), a post-processing framework for enhancing Unsupervised Video Object Segmentation (UVOS), particularly in multi-object scenarios. Our approach improves segmentation by prioritizing frequently appearing objects while filtering out noise and refining temporal consistency through a three-stage correction method.
Applying FTIO to DEVA resulted in state-of-the-art performance on the Unsupervised DAVIS 2017 dataset, demonstrating its effectiveness in object selection and temporal stability.

Future work could extend FTIO to more diverse datasets and integrate it with deep learning architectures to further enhance segmentation consistency and robustness.
\section{Supplementary Materials}
\label{sec:supplementary}

\begin{figure*}[hbt!]
\centering
  \includegraphics[width=\textwidth]{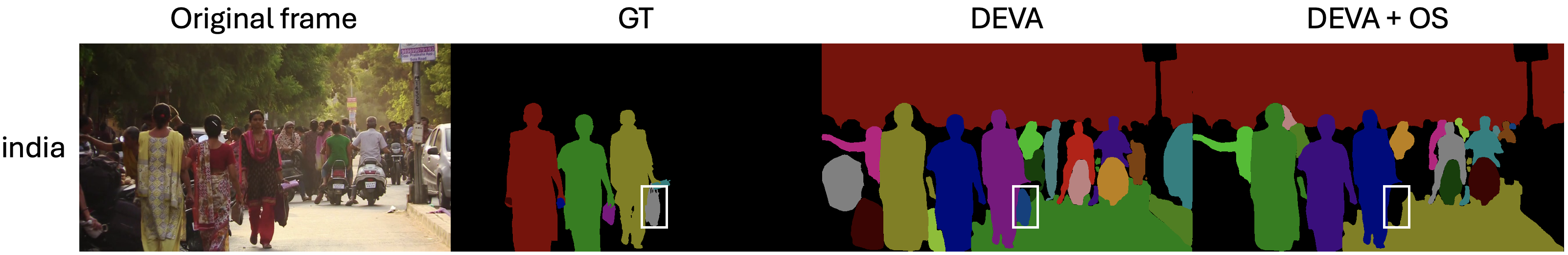}
  \caption[Missing object in FTIO (OS) predictions that was predicted correctly in DEVA~\cite{Cheng01a}]{Missing object in FTIO (OS) predictions that was predicted correctly in DEVA~\cite{Cheng01a}. Considering the object masks inside the white boxes, the bag with a gray mask in the ground truth of the ``india'' video sequence (that is predicted by DEVA~\cite{Cheng01a}) is missed when using OS post-processing. The ``DEVA'' and ``DEVA + OS'' images are obtained by reproducing the results of DEVA~\cite{Cheng01a} and applying OS phase of the FTIO approach to DEVA~\cite{Cheng01a}, respectively, using the code provided in the Github repository of DEVA~\cite{Cheng01a}~\cite{DEVA_Github} (licensed under CC BY-NC-SA 4.0). The source of the ``Original frame'' and ``GT'' images on the left side of this figure is from the multi-object unsupervised DAVIS dataset~\cite{Perazzi_2016_CVPR,Pont-Tuset01a,Caelles01a} (licensed under CC BY-NC 4.0). The white boxes are added manually to specify the object(s) of interest in each image.}
  \vspace{+1em}
  \label{fig:os_failure}
\end{figure*}

\begin{figure*}[hbt!]
\centering
  \includegraphics[width=\textwidth]{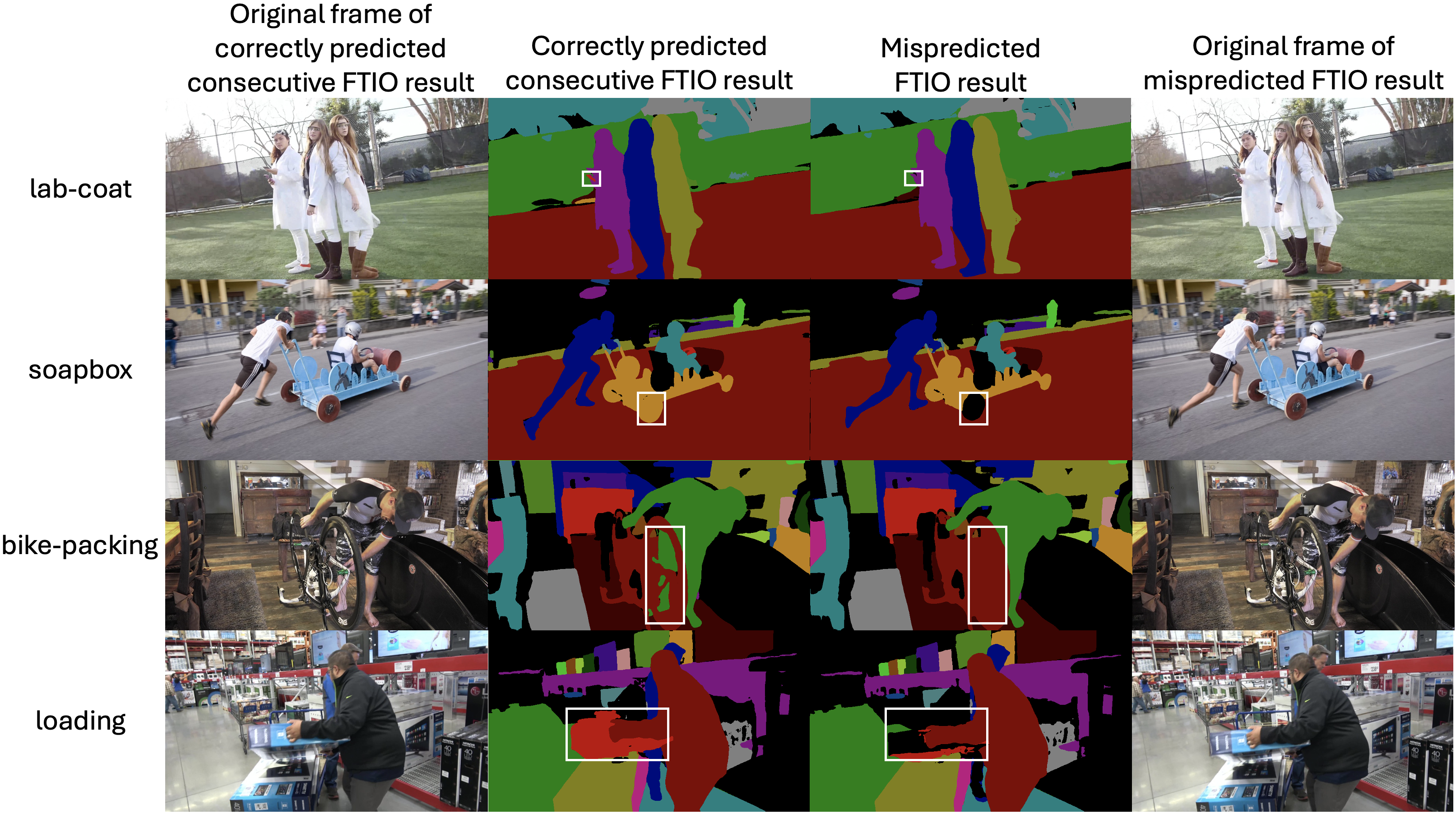}
  \caption[Frames with masks that remain inconsistent (even after applying TC)]{Frames with masks that remain inconsistent (even after applying TC). Considering the object masks inside the white boxes, when the object masks are tiny (``lab-coat''), or the relative size of inconsistent parts in those masks is considerably small (``soapbox''), the propagation of temporal consistency stops. When the intensity of pixels in inconsistent parts of object masks is highly similar to other objects in its neighborhood (``bike-packing''), or when the masks have some errors and include parts of other objects with different intensity of pixels (``loading''), the existing inconsistencies may remain. The images in the ``Correctly predicted consecutive FTIO result'' and ``Mispredicted FTIO result'' columns are some results of applying the FTIO approach to the code which is provided in the Github repository of DEVA~\cite{Cheng01a}~\cite{DEVA_Github} (licensed under CC BY-NC-SA 4.0). The source of the images in the ``Original frame of correctly predicted consecutive FTIO result'' and ``Original frame of mispredicted FTIO result'' columns is from the DAVIS dataset~\cite{Perazzi_2016_CVPR} (licensed under CC BY-NC 4.0). The white boxes are added manually to specify the object(s) of interest in each image.}
  \vspace{+1em}
  \label{fig:tc_failure}
\end{figure*}

\begin{table}[hbt!]
  \centering
  \caption[Comparison of inference time per frame (in seconds) between DEVA~\cite{Cheng01a} and FTIO]{Comparison of inference time per frame (in seconds) between DEVA~\cite{Cheng01a} and FTIO. It should be noted that the provided inference time per frame for both FTIO (OS) and FTIO only includes the post-processing steps.}
  \vspace{+1.5em}
  \begin{tabular}{p{5em}|cc}
    \hline
    {\textbf{Method}} & {\textbf{Validation}} & {\textbf{Test-dev}}\\
    \hline
    DEVA~\cite{Cheng01a} & 0.31 & 0.45\\
    FTIO (OS) & 0.05 & 0.05\\
    FTIO & 6.77 & 7.10\\
    \bottomrule
  \end{tabular}
  \label{tab:inference_time_comparison}
\end{table}

\newpage





\bibliography{ecai-paper}

\end{document}